\documentclass[letterpaper, 10 pt, conference]{ieeeconf}  % Comment this line out
\IEEEoverridecommandlockouts                              % This command is only
\usepackage{graphics} % for pdf, bitmapped graphics files
\usepackage{epsfig} % for postscript graphics files
\usepackage{mathptmx} % assumes new font selection scheme installed
\usepackage{times} % assumes new font selection scheme installed
\usepackage{amsmath} % assumes amsmath package installed
\usepackage{amssymb}  % assumes amsmath package installed
\usepackage{multirow}
\def\inv{\vspace*{-0.15cm}}

\usepackage[usenames, dvipsnames]{color}

\title{\LARGE \bf Segmentation of Overlapped Steatosis in Whole-Slide Liver Histopathology Microscopy Images}

\author{Mousumi Roy, Fusheng Wang, George Teodoro, Miriam B Vos, Alton Brad Farris, and Jun Kong
\thanks{Mousumi Roy and Fusheng Wang are with the Stony Brook University, Dept. of Computer Science, Stony Brook, NY 11794 ({\tt\footnotesize \{mousumi.roy, fusheng.wang\}@stonybrook.edu});  George Teodoro is with the University of Bras\'{\i}lia, Dept. of Computer Science,  Bras\'{\i}lia, DF, Brazil ({\tt\footnotesize glmteodoro@gmail.com});  Miriam B Vos, Alton Brad Farris, and Jun Kong are with the Emory University,  Atlanta, GA 30322 ({\tt\footnotesize \{mvos, abfarri, jun.kong\}@emory.edu}); The studies involving human subjects were approved by the Emory University IRB.}
}

\begin{document}

\maketitle
\thispagestyle{empty}
\pagestyle{empty}

%%%%%%%%%%%%%%%%%%%%%%%%%%%%%%%%%%%%%%%%%%%%%%%%%%%%%%%%%%%%%%%%%%%%%%%%%%%%%%%%
\begin{abstract}
An accurate steatosis quantification with pathology tissue samples is of high clinical importance. However, such pathology measurement is manually made in most clinical practices, subject to severe reader variability due to large sampling bias and poor reproducibility. Although some computerized automated methods are developed to quantify the steatosis regions, they present limited analysis capacity for high resolution whole-slide microscopy images and accurate overlapped steatosis division. In this paper, we propose a method that extracts an individual whole tissue piece at high resolution with minimum background area by estimating tissue bounding box and rotation angle. This is followed by the segmentation and segregation of steatosis regions with high curvature point detection and an ellipse fitting quality assessment method. We validate our method with isolated and overlapped steatosis regions in liver tissue images of 11 patients. The experimental results suggest that our method is promising for enhanced support of steatosis quantization during the pathology review for liver disease treatment.
\end{abstract}
%%%%%%%%%%%%%%%%%%%%%%%%%%%%%%%%%%%%%%%%%%%%%%%%%%%%%%%%%%%%%%%%%%%%%%%%%%%%%%%%
\section{INTRODUCTION}
An excessive accumulation of lipid in liver cells results in liver steatosis~\cite{Tiniakos}. Among all possible causes of this disease, alcohol, obesity, and type II diabetes mellitus, are known as common factors that contribute to steatosis formation~\cite{Brunt2010}. Steatosis is a defining feature of nonalcoholic fatty liver disease which may also include inflammation, hepatocyte ballooning and fibrosis and eventually cirrhosis~\cite{Tiniakos}. Steatosis is assessed in a liver biopsy using a number of grading systems in order to determine the severity of involvement of the liver by fat. For example, the Brunt system has been used in a number of clinical studies to determine steatosis and steatohepatitis (NASH) severity, notably by the NASH Clinical Research Network (NASH CRN)~\cite{Kleiner, Neuschwander}. 

Quantification of the steatosis component is important in many clinical scenarios, especially in liver transplantation~\cite{Nocito}. It has been unveiled that liver transplant recipients receiving livers rich in steatosis tend to have a higher rate of primary graft dysfunction and/or renal failure~\cite{Feng}. In practice, however, there is a lack of objective ways to measure steatosis degree, leading to a severe reader variability~\cite{Marsman2004}. Although some computer based analysis methods have been developed, they cannot be deployed for clinical practice, as they cannot accommodate whole-slide image analysis and have worse analysis results when steatosis components are overlapped in large tissue areas~\cite{Marsman2004, Liquori, Kong2011}. 

To address this problem, we propose a new image analysis method that can segregate clumped steatosis in pairs in whole-slide tissue histopathology images. The proposed method consists of whole tissue component extraction, steatosis detection, and overlapped steatosis component segregation. The paper is organized as follows. We discuss our algorithm in Section~\ref{sec:method}. Experimental results are demonstrated in Section~\ref{sec:result}. In Section~\ref{sec:conclusion}, we conclude our paper.

\section{METHODOLOGY~\label{sec:method}}
The developed algorithm for steatosis analysis consists of a sequence of steps. First, images containing whole tissue components are automatically detected and extracted from whole-slide microscopy images at highest resolution. To minimize the non-tissue areas, we transform the resulting images by rotation and interpolation. To reduce memory usage, we use two image resolutions. We detect tissue regions and estimate rotation angles at the low resolution, whereas we extract whole tissue components from whole-slide images at the high resolution. This is followed by the segregation of steatosis regions in each tissue piece by high curvature point detection and an ellipse fitting quality assessment method. Each step consists of a sequence of processing modules that are presented in details in the following space.

\subsection{Tissue Extraction from Whole Slide Images}
\begin{figure}[tb!]
\centerline{
\includegraphics[width=\linewidth]{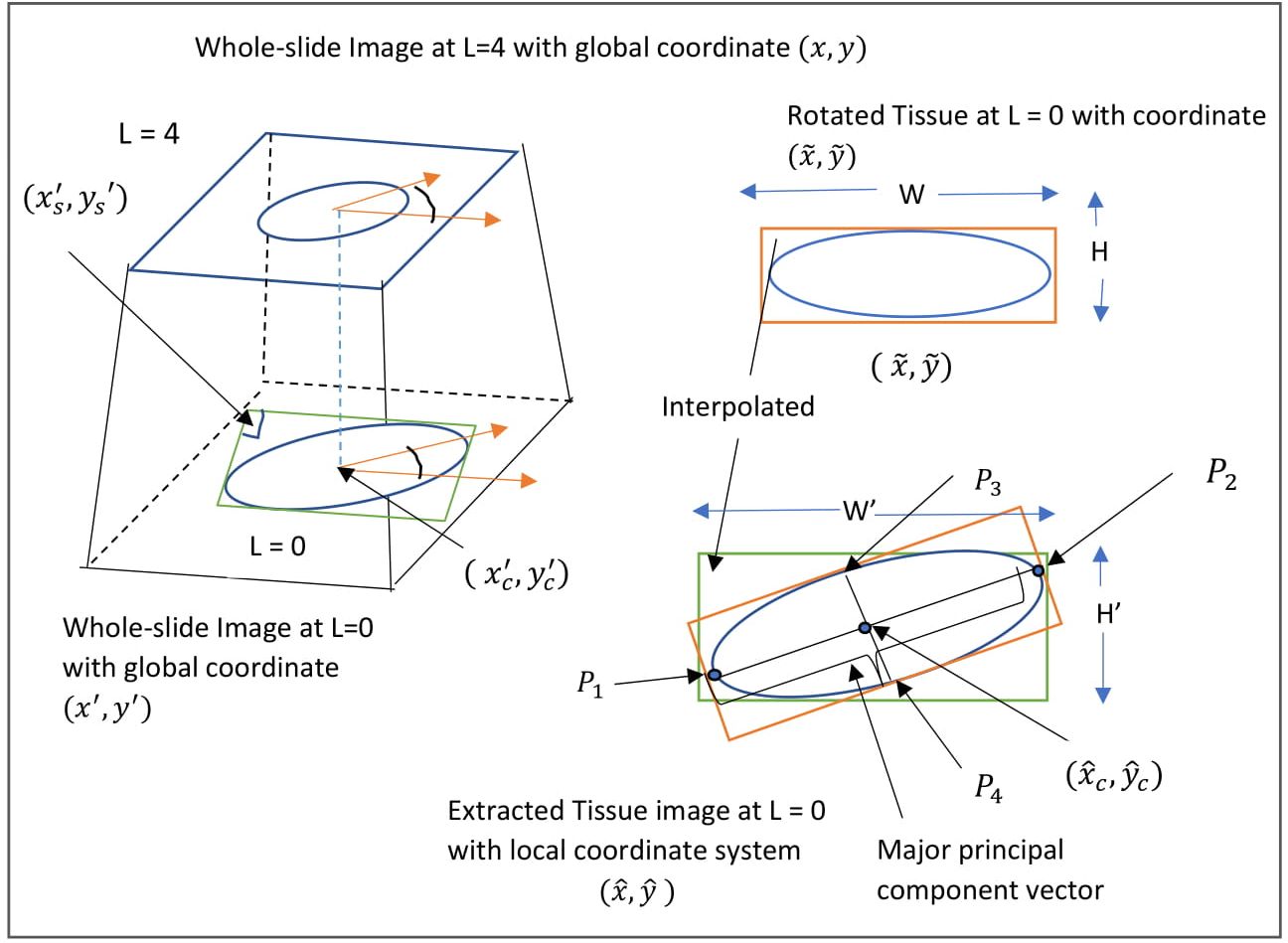}
} \inv\inv\inv
\caption{The overall schema of whole tissue component extraction at $L = 0$ is presented.}\label{fig:draw} \inv\inv%\inv
\end{figure}

The first stage of our analysis focuses on extracting each complete tissue component with minimum background area at the highest image resolution, i.e. $L = 0$, where $L$ is the image resolution level. In order to estimate the spatial bound of each whole tissue component, we call the OpenSlide API~\cite{openslide} and extract a whole-slide image representation at a lower image resolution ($L=4$). One such image is demonstrated in Fig.~\ref{fig:draw}. Aiming at extracting each tissue component with minimum background area, we identify each tissue component mask by Otsu's thresholding~\cite{Otsu} and compute the rotation angle using Principal Component Analysis (PCA) of the resulting single tissue mask~\cite{PCA}.
 
Each resulting image containing one whole tissue component at $L = 4$ is processed further to find the tissue boundary points at $L = 4$. A minimum bounding rectangle around each complete tissue component at $L = 4$ is fitted to specify the Region of Interest(ROI) in a whole-slide microscopy image. The bounding box of the ROI at $L = 0$ can be derived from the same at $L = 4$ by the following transformation:
$
\begin{pmatrix}
x'\\  y'
\end{pmatrix}
= 2^L
\begin{pmatrix}
x\\  y
\end{pmatrix}
$
% \begin{equation}~\label{eqn:low2high}
% \begin{pmatrix}
% x'\\  y'
% \end{pmatrix}
% = 2^L
% \begin{pmatrix}
% x\\  y
% \end{pmatrix}
% \end{equation}
\noindent where $(x, y)^T$ and $(x', y')^T$ are pixel coordinates in the whole-slide image at $L = 4$ and $L = 0$, respectively.

Similarly, the tissue boundary points at $L = 0$ can be derived from those at $L = 4$ by the same mapping. In this way, we can readily extract only the tissue ROIs at $L = 0$ from a raw whole-slide microscopy image. Let us denote $(\hat{x}, \hat{y})^T$ as the local pixel coordinates with respect to the extracted tissue image at $L = 0$, and $(x', y')^T$ as the global coordinates for the whole-slide image at $L = 0$. The tissue boundary pixels in local coordinate system $(\hat{x}, \hat{y})^T$ can be obtained from the global coordinates $(x', y')^T$ by the following transformation: 
\begin{equation}~\label{eqn:transformation}
\begin{pmatrix}
x'\\  y'
\end{pmatrix}
=
\begin{pmatrix}
\hat{x}\\ \hat{y} 
\end{pmatrix}
+
\begin{pmatrix}
 x_c' \\ y_c' 
\end{pmatrix}
-
\begin{pmatrix}
 W'/2 \\ H'/2
\end{pmatrix}
\end{equation}
\noindent where $(x_c', y_c')^T$ is the center of the mapped bounding box at $L = 0$ in the global coordinate system;  $W'$ and $H'$ are the width and height of the bounding box around the tissue ROIs at $L = 0$.

Next, we aim to generate the rotated tissue at $L = 0$ so that the resulting image contains minimum background area. We denote the width and height of the bounding box of the rotated tissue image at $L = 0$ as $W$ and $H$, respectively. To find $W$ and $H$, we compute the center of the extracted tissue image $(\hat{x_c}, \hat{y_c})^T$ with respect to the local coordinate system. We compute $(\hat{x_c}, \hat{y_c})^T$ by subtracting the top left coordinate $(x_s', y_s')^T$ from the center $(x_c', y_c')^T$ of the mapped bounding box at $L = 0$ in the global coordinate system as shown in Fig.~\ref{fig:draw}. Since the rotation angle at different image resolutions (i.e. $L = 4$ and $L = 0$) remains same, we know the equations of lines in parallel to major and minor principal component vectors that pass through the center $(\hat{x}_c, \hat{y}_c)^T$ at $L = 0$. In this way, we can detect tissue boundary point pairs $(P_1, P_2)$ and $(P3, P4)$ that are on these two lines, as depicted in Fig.~\ref{fig:draw}. The resulting distances between the tissue boundary point pairs $(P1, P2)$ and $(P3, P4)$ on lines in parallel to major and minor principal component vectors are width $W$ and height $H$ of the bounding box of the rotated tissue at $L = 0$. From the estimated width and height, we create a mesh grid of size $W \times H$ representing the bounding box of the rotated tissue image at $L = 0$. 

Let us denote $(\tilde{x}, \tilde{y})^T$ as the coordinate system for the rotated tissue image at $L = 0$. With the following transformation, we map each pixel $(\tilde{x}, \tilde{y})^T$ in the rotated tissue image at $L=0$ to a corresponding pixel $(\hat{x}, \hat{y})^T$ in the non-rotated tissue image at $L = 0$:
\begin{equation}
\begin{pmatrix}
\hat{x}\\ \hat{y} 
\end{pmatrix}
=
R^{-1} \left( 
\begin{pmatrix}
\tilde{x}\\  \tilde{y}
\end{pmatrix}
-
\begin{pmatrix}
 W/2 \\ H/2
\end{pmatrix}
\right)
+
\begin{pmatrix}
 W'/2 \\ H'/2
\end{pmatrix}
\end{equation}
\noindent where $R$ is a rotation matrix with rotational angle found by Principal Component Analysis (PCA) of the single tissue mask at $L = 4$.

If the mapped pixel $(\hat{x}, \hat{y})^T$ is outbound of the non-rotated tissue image domain, the corresponding pixel $(\tilde{x}, \tilde{y})^T$ is assigned a user defined color value. For any pixel $(\tilde{x}, \tilde{y})^T$ that is mapped within the non-rotated tissue image domain, we interpolate \cite{Recipe} its color value with the color information of the neighboring pixels in the non-rotated tissue image. The resulting interpolated image after this process contains a rotated complete tissue with minimum non-tissue region at $L = 0$. Such images are further processed for steatosis identification and segregation.

\subsection{Quantification of Steatosis Components}
The rotated tissues at $L = 0$ are analyzed further to identify steatosis regions. We notice that the steatosis components are present in varying shapes within a single tissue part. Individual steatosis components are mostly circular in shape whereas the clumped steatosis regions have irregular shapes due to multiple steatosis overlap. Additionally, we notice that some non-steatosis regions present similar intensities to isolated and overlapped steatosis instances, making it challenging to detect true steatosis regions accurately.

To address this problem, we first convert each rotated tissue image to gray scale and enhance its contrast with adaptive histogram equalization~\cite{Karel} method. Next, we use the hysteresis thresholding to binarize the gray scale image with low and high threshold values as $0.65$ and $0.8$, respectively~\cite{hysteresis}. For the remaining tissue regions, we use morphological operations to smooth the boundary of steatosis components and to remove small objects that are too small to be steatosis in the post-processing step.

As most steatosis regions are circular in shape, we compute the inverse of {\it Circularity} for each connected component within a tissue as: $C^{-1} =  P^2 / (4 \pi A) $ where, $C$, $A$ and $P$ are the circularity, area size, and perimeter length of a steatosis candidate, respectively. The non-steatosis regions with inverse {\it Circularity} value $(> 3)$ are mostly non-circular in shape and are removed from the detected steatosis candidate set. In the remaining set, there is still a large number of overlapped steatosis regions with irregular contours, making it difficult to segregate individual steatosis regions. Thus, we need to differentiate individual steatosis instances with regular shape and no touching neighbors from those candidates with irregular shape due to overlapped steatosis. We classify these two classes of instances by the {\it Solidity} measure derived from each connected component. {\it Solidity} is computed as the ratio of {\it Area} to {\it Convex Area}, representing the proportion of pixels in a steatosis region as compared to those in its convex hull. Those candidates with strong {\it Solidity} measure $(>0.95)$ are detected as single steatosis component with no overlapping neighbors. By contrast, the resulting candidates with {\it Solidity} measure below $0.95$ can be either non-steatosis components or overlapped steatosis instances. Therefore, we next measure the {\it Extent} property computed as the ratio of the number of pixels in the candidate region to that in the  bounding box. We remove those with {\it Extent} value $< 0.5$, as they represent non-steatosis objects. The remaining instances with strong {\it Extent} measure $(>= 0.5)$ are considered as overlapped steatosis.

In our study, we notice that it is necessary to consider all these properties of steatosis candidate regions (i.e. {\it Circularity}, {\it Solidity} and {\it Extent}) to identify the single non-touching and the overlapped steatosis instances accurately. Dropping any one of these properties results in loss of a large number of overlapped steatosis regions, leading to erroneous steatosis quantification results. To segregate such clumped steatosis regions, we propose a methodology to produce the dividing borders by an ellipse fitting assessment procedure.

\subsection{Segregation of Overlapped Steatosis regions}
\begin{figure}[tb!]
\centerline{
\includegraphics[width=\linewidth]{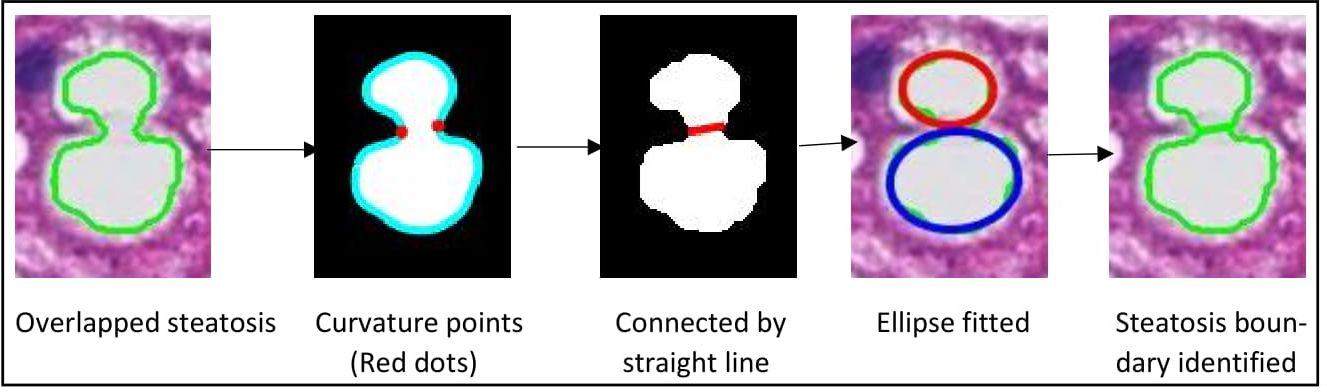}
} \inv \inv \inv
\caption{Overlapped steatosis is assessed by the ellipse fitting quality.}\label{fig:segregation} \inv \inv \inv
\end{figure}

As the vast majority of steatosis overlapped regions invovle only two steatosis components, we aim to address the two steatosis overlapping problem specifically. We illustrate the segregation analysis process in Fig.~\ref{fig:segregation}. 

For each overlapped steatosis region, we identify high curvature points by $\kappa=\frac{{x}'{y}''-{y}'{x}''}{({x}'^{2}+{y}'^{2})^{3/2}}$, where $x$ and $y$ are coordinates of the steatosis boundary points~\cite{curvature}. For a long concave contour segment, we merge adjacent curvature points to one aggregated point representation. For each curvature point $p_i$ in a detected high curvature point set $P=\{p_i | i = 1, 2, \cdots, N\}$, it is joined by a straight line with every other point $p_j$ for division assessment. For each such candidate line, the overlapped steatosis component is partitioned into two sub-regions. We assess the partitioning quality by fitting an ellipse~\cite{Book} to each sub-region. The formulated fitting quality measure $F(i, j)$ is defined as the ratio of the area intersected by the fitting ellipse and the steatosis sub-region to their union area. With such $F(i, j)$ defined, we connect such $(p_i, p_j) \in P$ with a dividing line when the resulting $F(i, j)$ of the two fitting sub-regions is sufficiently strong $(> 0.7)$ and maximum of all $F(i, j)$ measures associated with all possible pairs of points from set $P$. If the best fitting quality $F(i, j)$ is not strong enough, the associated candidate region is considered as non-separable. 

\section{Experimental results~\label{sec:result}}
Our dataset consists of whole slide pathology microscopy images of human liver biopsies. All histologic tissue slides are stained with hematoxylin and eosin before they are scanned and converted to whole-slide digital images. The resolution of resulting whole-slide microscopy images are overwhelmingly large ($30,000\times 20,000$ pixels on average). 

\begin{figure}[tb!]
\centerline{
\includegraphics[width=\linewidth]{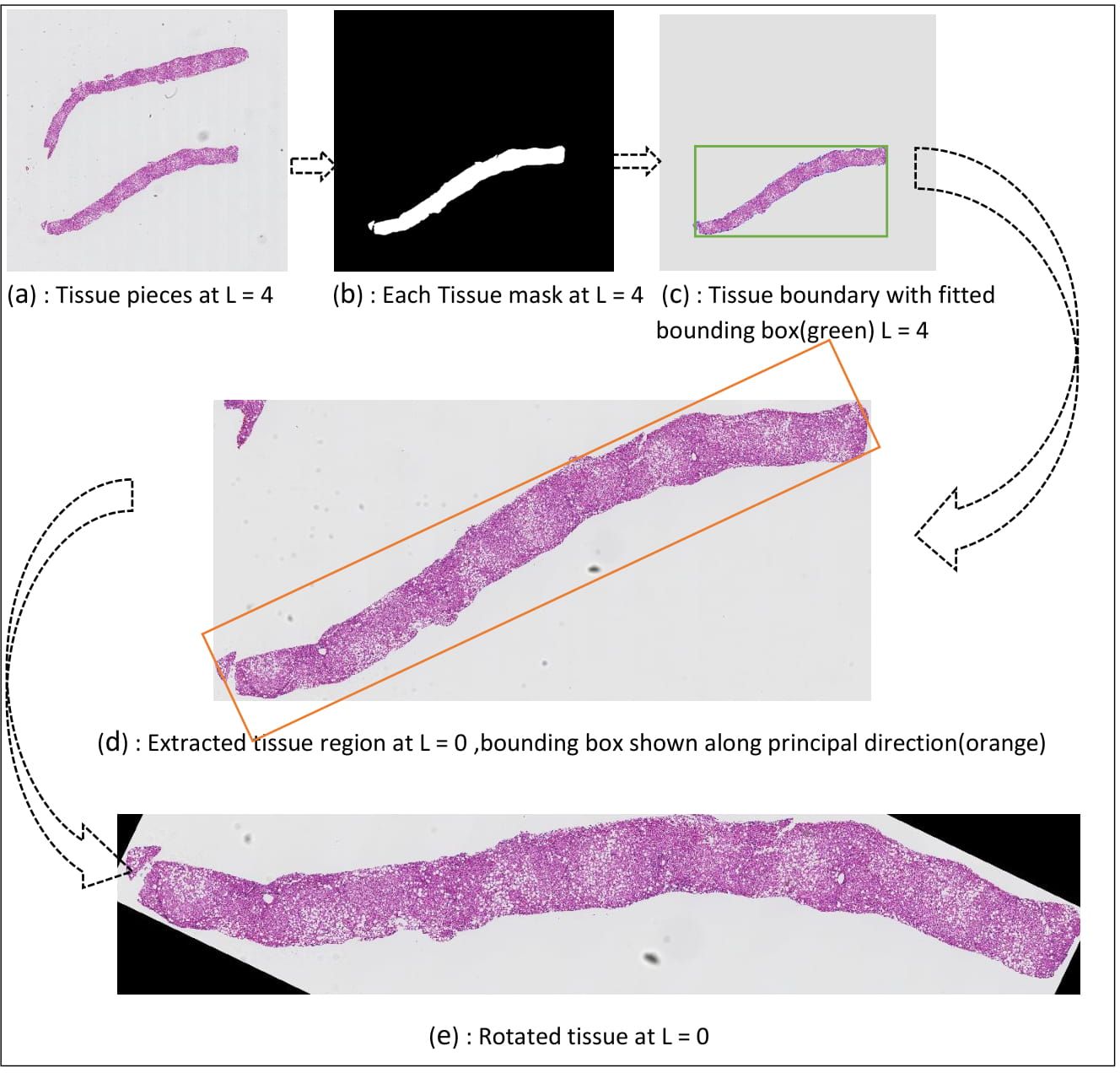}
} \inv %\inv \inv
\caption{Intermediate results of the whole tissue extraction across different resolution levels are demonstrated.}\label{fig:rotation} \inv%\inv\inv
\end{figure}

%We present the intermediate results of the whole tissue extraction method in Fig.~\ref{fig:rotation}. Specifically, 
We demonstrate a typical whole-slide image containing multiple tissue pieces at an intermediate resolution $L=4$ in Fig.~\ref{fig:rotation}(a). Each complete tissue component is analyzed independently. In Fig.~\ref{fig:rotation}(b)-(e), we present a detected tissue mask at $L=4$ with Otsu's thresholding, the corresponding tissue bounding box at $L=4$, the extracted whole tissue image at $L=0$, and finally the rotated tissue image with minimum background area at $L=0$, respectively. 

Due to the overwhelmingly large image resolution, it is challenging to extract the whole tissue piece from whole-slide microscopy images at a high resolution. Most existing studies partition whole-slide microscopy images into patches and execute the analysis on each patch separately. As a result, steatosis components on the borders of such patch can be erroneously analyzed. To overcome this shortcoming, our method extracts whole tissue image at the highest resolution with minimum background area. With such extracted images, we are able to identify steatosis regions more accurately without having to analyze steatosis on patch borders. 

\begin{figure}[bt!]
\centerline{
\includegraphics[width=\linewidth]{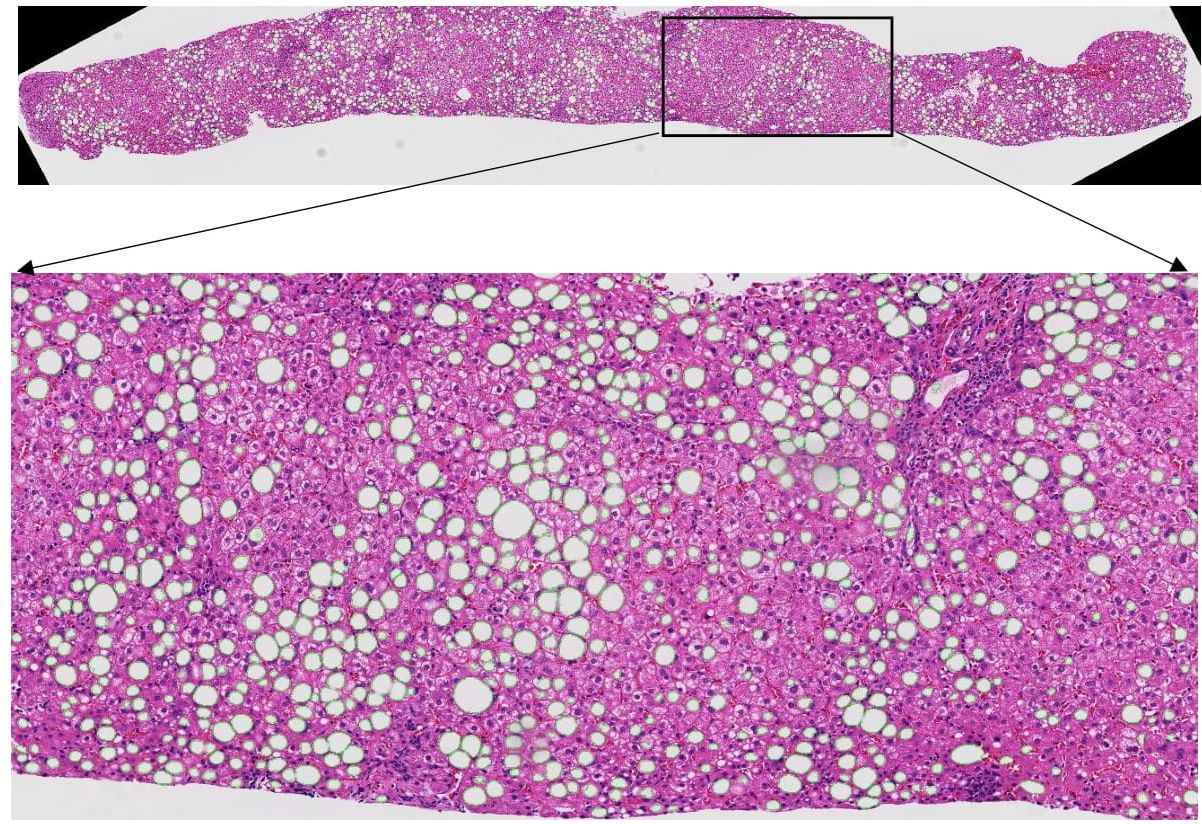}
} \inv \inv %\inv
\caption{Steatosis regions from a typical whole tissue piece at $L=0$ are segmented and segregated.}\label{fig:whole} \inv \inv \inv \inv
\end{figure}

\begin{figure}[bth!]
\centerline{
\includegraphics[width=\linewidth, height=6.5cm]{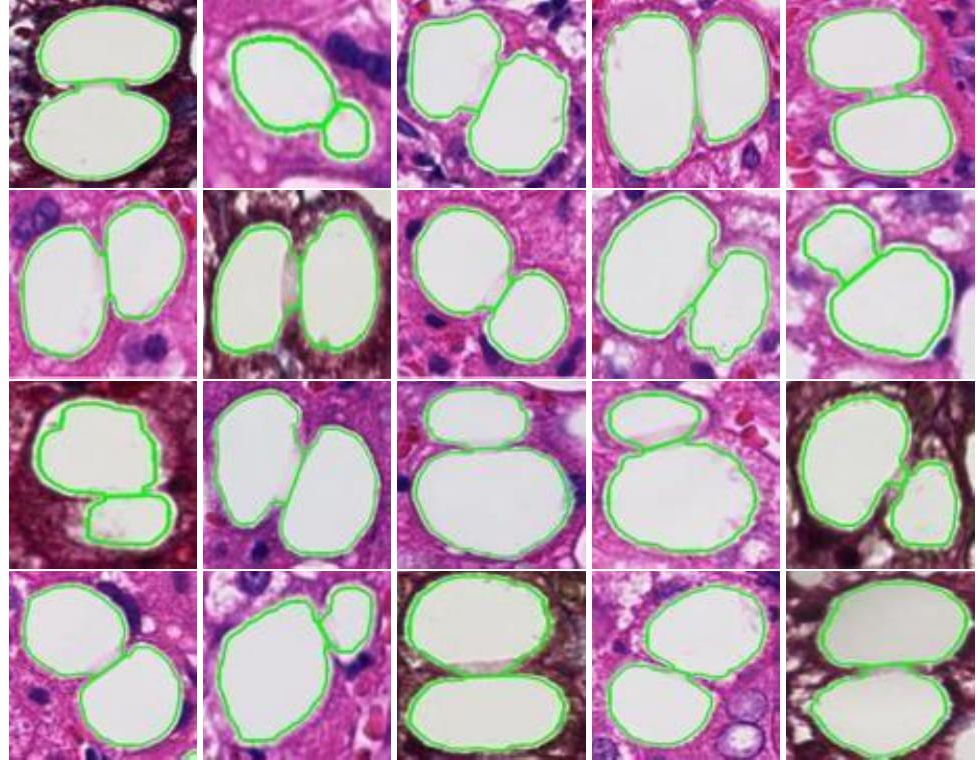}   %width=70mm,scale=0.5
} \inv \inv
\caption{A panel of segregated steatosis components is presented.}\label{fig:panel} \inv\inv%\inv
\end{figure}

For the validation purpose, we test our method with liver whole slide microscopy images of 11 patients. In general, there are thousands of Isolated Steatosis (IS) regions detected from each whole-slide image, as depicted in Fig.~\ref{fig:whole}. By contrast, we observe a large variation in number of Overlapped Steatosis (OS) regions among these slides. In our test, we analyze tissue pieces with (1) abundant, (2) few, and (2) absent overlapped steatosis regions. Due to the overwhelmingly large number of steatosis regions, we randomly select 100 validation instances from isolated and another 100 from overlapped steatosis result sets for each whole slide image, respectively. When the number of overlapped steatosis regions from a slide is less than 100, we include them all for validation. We present the quantitative evaluation results and the total number (round off value to hundreds) of detected instances of isolated and overlapped steatosis cases in Table~\ref{table_example}. In addition, we present a panel of representative steatosis segregation results in Fig.~\ref{fig:panel}. %Overall, the accuracy of our proposed method for steatosis segmentation and segregation is promising. 
With our dataset, the current parameter settings result in robust steatosis segmentation. Almost all isolated steatosis instances in the validation dataset are correctly segmented for all patients. Although our segregation method fails in some cases, the overall segregation performance is satisfying. Of those problematic instances, %where our method fails%, 
we find that they are either clumped cases with more than two steatosis components or over-segmented steatosis regions. In future, we will improve this method to accommodate more than two overlapped steatosis components.

\begin{table}[h]
\small
\centering
\caption{Segmentation accuracy and the validation number (in parenthesis) for Isolated Steatosis (IS) and Overlapped Steatosis (OS) from 11 patients (indicated by PID) with abundant, low, and absent overlapped steatosis regions. %Numbers in bracket indicate the total number of instances in validation set.
}~\label{table_example}
\inv
 \begin{tabular}{| c| c | c c | c c|} 
 \hline
\multirow{2}{*}{OS} & \multirow{2}{*}{PID} & \multicolumn{2}{|c|}{Total Detection $\#$} & \multicolumn{2}{|c|}{Validation Accuracy (Total $\#$)} \\
 \cline{3-4} \cline{5-6}
 &  & IS & OS & IS & OS \\ 
\hline\hline
\multirow{5}{*}{\rotatebox[origin=c]{90}{Abundant}} & P01 & 2000 & 300 & 100$\%$, (100) & 87$\%$, (100) \\ 
\cline{2-6}
 & P02 & 2500 & 600 & 100$\%$, (100) & 92$\%$, (100) \\
\cline{2-6}
& P03 & 1500 & 450 & 100$\%$, (100) & 99$\%$, (100) \\ 
\cline{2-6}
& P04 & 3500 & 500 & 100$\%$, (100) & 99$\%$, (100) \\ 
\cline{2-6}
& P05 & 1600 & 200 & 99$\%$, (100) & 96$\%$, (100) \\ 
\hline
\multirow{4}{*}{\rotatebox[origin=c]{90}{Low}} & P06 & 800 & 19 & 100$\%$, (100) & 100$\%$, (19) \\ 
\cline{2-6}
&  P07 & 2100 & 19 & 100$\%$, (100) & 89$\%$, (19) \\ 
\cline{2-6}
&  P08 & 1500 & 7 & 100$\%$, (100) & 86$\%$, (7) \\ 
\cline{2-6}
& P09 & 1200 & 20 & 99$\%$, (100) & 100$\%$, (20) \\
\hline
\multirow{2}{*}{\rotatebox[origin=c]{90}{Absent}} & P10 & 400 & - & 99$\%$, (100) & - \\  [0.1cm]
\cline{2-6}
&  P11 & 900 & - & 100\%, (100) & - \\[0.1cm]
      \hline
\end{tabular}\inv
\end{table}

\section{CONCLUSIONS~\label{sec:conclusion}}
In this paper, we propose a new image analysis method to segregate clumped steatosis components in whole-slide liver tissue microscopy images. The developed method consists of whole tissue image extraction, steatosis detection, and overlapped steatosis segregation. We validate our method with a large number of isolated and overlapped steatosis cases. The testing results suggest that our method is promising for enhanced support of steatosis quantization during the pathology review for liver disease treatment.


\begin{thebibliography}{99}
{\scriptsize
\bibitem{Tiniakos} Tiniakos DG, Vos MB, Brunt EM. Nonalcoholic fatty liver disease: pathology and pathogenesis. Annual review of pathology.5:145-171, 2010.
\bibitem{Brunt2010} Brunt EM. Pathology of nonalcoholic fatty liver disease. Nature reviews. Gastroenterology \& hepatology. 7(4):195-203, Apr 2010.
\bibitem{Kleiner} Kleiner DE, Brunt EM, Van Natta M, et al. Design and validation of a histological scoring system for nonalcoholic fatty liver disease. Hepatology. 41(6):1313-1321, Jun 2005.
\bibitem{Neuschwander} Neuschwander-Tetri BA, Clark JM, Bass NM, et al. Clinical, laboratory and histological associations in adults with nonalcoholic fatty liver disease. Hepatology. 52(3):913-924, Sep 2010.
\bibitem{Nocito} Nocito A, El-Badry AM, Clavien PA. When is steatosis too much for transplantation? Journal of hepatology. 45(4):494-499, Oct 2006.
\bibitem{Feng} Feng S. Steatotic livers for liver transplantation--life-saving but at a cost. Nature clinical practice. Gastroenterology \& hepatology. 5(7):360-361, Jul 2008.
\bibitem{Marsman2004} Marsman H, Matsushita T, et al., Assessment of donor liver steatosis: pathologist or automated software? Human pathology. 35(4):430-435,2004.
\bibitem{Liquori} Liquori GE, Calamita G, Cascella D, Mastrodonato M, Portincasa P, Ferri D. An innovative methodology for the automated morphometric and quantitative estimation of liver steatosis. Histol Histopathol. 24(1):49-60, Jan 2009.
\bibitem{Kong2011} Kong J, Lee M, Bagci P, Sharma P, Martin D, Adsay NV, Saltz J, Farris A. Computer-based Image Analysis of Liver Steatosis with Large-scale Microscopy Imagery and Correlation with Magnetic Resonance Imaging Lipid Analysis, IEEE Conference of bioinformatics and biomedicine, pp.333-338, 2011.
\bibitem{openslide} Goode A, Gilbert B, Harkes J, Jukic D, Satyanarayanan M. OpenSlide: A Vendor-Neutral Software Foundation for Digital Pathology, Journal of Pathology Informatics, 4:27,  2013.
\bibitem{PCA} Abdi, H., Williams, L. J. (2010). Principal component analysis. Wiley Interdisciplinary Reviews Computational Statistics, 2(4), 433–459.
\bibitem{Auger}Auger J., Schoevaert D. and Martin E.D. Comparative study of automated morphometric and semiquantitative estimations of alcoholic liver steatosis. Anal. Quant. Cytol. Histol. 8, 56-62, 1986. 
\bibitem{Karel}Karel Z., “Contrast Limited Adaptive Histograph Equalization.” Graphic Gems IV. San Diego: Academic Press Professional, 474–485, 1994. 
\bibitem{Otsu} Otsu, N., A Threshold Selection Method from Gray-Level Histograms. IEEE Transactions on Systems, Man, and Cybernetics, Vol. 9, No. 1,pp. 62-66,  1979.
\bibitem{hysteresis} Yu Y., Li Z., Liu B., Liu X., An Adaptive Unimodal and Hysteresis Thresholding Method. Communications in Computer and Information Science, vol 472. Springer, Berlin, Heidelberg,2014.
\bibitem{curvature} Wen Q., Chang H., and Parvin, B., A Delaunay triangulation approach for segmenting clumps of nuclei. IEEE International Symposium on Biomedical Imaging: From Nano to Macro, pp.9-12, 2009.
\bibitem{Book}Bookstein F.L., Fitting Conic Sections to Scattered Data, Computer Graphics and Image Processing, pp. 56-71, 1979.
\bibitem{Recipe}Numerical Recipes in C, Cambridge University Press, pp.123-128, 1988-92.%ISBN 0-521-43108-5.
}
\end{thebibliography}
\end{document}